%% file: MainSubmitted.tex
\documentclass[letterpaper, 10 pt, conference]{ieeeconf}
\IEEEoverridecommandlockouts                              

\usepackage{dblfloatfix}
\usepackage{epsfig}
\usepackage{psfrag,graphicx,color}
\usepackage[normalem]{ulem}
\usepackage{amsmath,amsfonts}
\usepackage{hyperref}
\usepackage{epsfig,psfrag,graphicx,color}
\usepackage{amsthm}
\usepackage{amssymb}
\usepackage{algpseudocode}
\usepackage{algorithm}
\usepackage{graphicx}
\usepackage{url} 
\usepackage{cleveref}
\input{bmit2e}

\input{mymacro}
\input MATHSYM.TEX

\newtheorem{problem}{Problem}

\usepackage{amssymb}  
\usepackage[table,xcdraw]{xcolor}
\def\tasym{\mathcal{T}}
\def\ta{_{\tasym}}
\def\sta#1{_{\tasym,#1}}
\def\hn{\bfzeta}
\def\p{P}

\DeclareMathOperator*{\argmin}{arg\,min}
\newcommand{\dispdot}[2][5mu]{\dot{#2\mkern#1}\mkern-#1}
\newcommand{\dispbar}[2][5mu]{\bar{#2\mkern#1}\mkern-#1}
\Crefname{equation}{Eq.}{Eqs.}
\Crefname{figure}{Fig.}{Figs.}
\Crefname{tabular}{Tab.}{Tabs.}
\Crefname{table}{Tab.}{Tabs.}
\Crefname{section}{Sec.}{Secs.}

\newlength{\Oldarrayrulewidth}

\usepackage{hhline}

\renewcommand\thesubsubsection{T\arabic{subsubsection}}

\makeatletter

\makeatletter
\IEEEtriggercmd{\reset@font\normalfont\fontsize{7.5pt}{8.40pt}\selectfont}
\makeatother
\IEEEtriggeratref{1}

\begin{document}

\title{\LARGE \bf  
A Data-Driven Approach for Contact Detection,  Classification and  Reaction in Physical Human-Robot Collaboration
}

\author{Martina Lippi${}^1$, Giuseppe Gillini${}^2$, Alessandro Marino${}^2$, Filippo Arrichiello${}^2$ 
\thanks{${}^1$ Roma Tre University, Italy, { martina.lippi@uniroma3.it} }
\thanks{{${}^2$University of Cassino and Southern Lazio, Italy
	 \{giuseppe.gillini, al.marino, f.arrichiello\}@unicas.it}}
  \thanks{*This work has been funded by the European Union’s Horizon 2020 research and innovation program under grant agreement No 101016906, project CANOPIES.}
  }
\maketitle

\begin{abstract}
This paper considers a scenario where a robot and a human operator share the same workspace, and the robot is able to both carry out autonomous tasks and physically interact with the human in order to achieve common goals. In this context, both intentional and accidental contacts between human and robot might occur due to the complexity of tasks and environment, to the uncertainty of human behavior, and to the typical lack of awareness of each other actions. 
Here, a two stage strategy based on Recurrent Neural Networks (RNNs) is designed to detect intentional and accidental contacts:  the occurrence of a contact with the human is detected at the first stage, while the classification between intentional and accidental is performed at the second stage.  An admittance control strategy or an evasive action is then performed by the robot, respectively. The approach also works in the case the robot simultaneously interacts with the human and the environment, where the interaction wrench of the latter is modeled via Gaussian Mixture Models (GMMs). Control Barrier Functions (CBFs) are included, at the control level, to guarantee the satisfaction of robot and task constraints while performing the proper interaction strategy. 
The approach has been validated on a real setup composed of a Kinova Jaco$2$ robot.
\end{abstract}

\section{Introduction}

Collaboration between humans and robots can envisage the case of pure workspace sharing, in which human and robot work on autonomous or coordinated tasks without the need of physical contact, or of voluntary exchange of forces to achieve a common goal. In both cases, the human safety is the primary issue. However, while in the workspace sharing case, the human safety can be achieved by delivering proper collision avoidance strategies that prevent unsafe contacts~\cite{Zanchettin_TRO2013,Lippi_TCST2020}, in the other case,  safety requirements are more challenging to  meet being the exchange of forces envisaged by the task itself (like in the case of kinesthetic teaching~\cite{kormushev2011imitation}). Indeed, in the latter case not only {\it intentional} contacts but also {\it accidental} collisions should be considered. 
The two cases obviously require the robot to adopt different reaction strategies. In particular, in the case of intentional contact, admittance or impedance control are traditionally employed, which confer a compliant behavior to the robot  structure through a mass-spring-damper model~\cite{Tsumugiwa_ICRA2002,Dimeas_TH2016}. 
On the contrary, accidental collisions are representative of unexpected and dangerous situations, which need to be promptly detected and handled by avoidance strategies like in~\cite{Zanchettin_TRO2013}. However, in order to undertake a suitable reaction, it is first required to detect the occurrence of a contact and, then, to recognize whether it is intentional or accidental~\cite{Haddadin_TRO2017}. 
Contact detection solutions often exploit the dynamical model of the robot and joint torque sensors, as in~\cite{DeLuca_ICRA2005}, or motor currents as in~\cite{Flacco_ICRA2013}. Other approaches, like the one in \cite{Kouris_RAL2018}, leverage the different frequency characteristics of accidental and intentional contacts to achieve classification in the frequency domain. Additional solutions might also leverage proper artificial skins for robots as in~\cite{Cheng_2019,Albini_IJRR2020}.

\begin{psfrags}
		\def\scal{0.9}  
		\def\scals{0.7} 
		\psfrag{Det}[cc][][\scal]{\shortstack[c]{\hspace{10pt}Detect and Recognize}}
		\psfrag{Rea}[cc][][\scal]{React}
		\psfrag{Int}[cc][][\scals]{Intentional}
		\psfrag{Acc}[cc][][\scals]{Accidental}
		\mytoppsfrag{6.3}{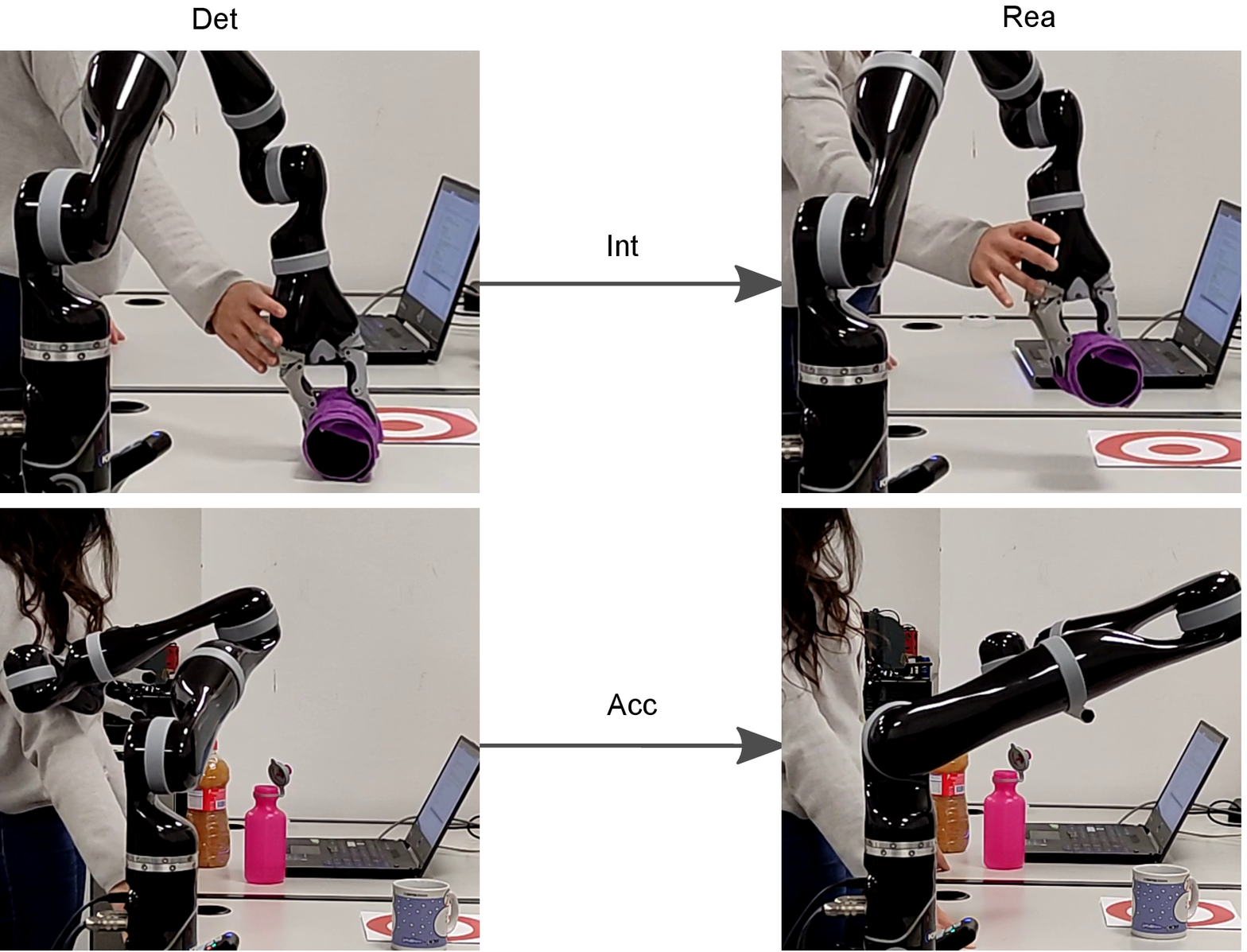}{-12pt}{  Framework example: the human contact is detected and classified (on the left) and the system reacts according to its nature (on the right), i.e. intentional (top row) or accidental (bottom row). }{fig:scene} 
	\end{psfrags}
	
Classical approaches, like the ones cited above, requires thresholds to be manually tuned in order to take into account sensor noise and different type of contacts, which results in poor flexibility and robustness of the overall system.
For this reason, data-driven approaches have been devised in recent years for contact detection and classification, due to their flexibility and capability of handling the non-linearity  and variety of the human-robot contact. In~\cite{Sharkawy_2018,sharkawy2020neural},  Neural Networks (NNs) are used to detect sole accidental collisions on the basis of data coming from joint torque sensors. The same objective is achieved in~\cite{heo2019collision}  by using a deep learning approach which requires the tuning of a moving time window. 
Finally, the study in~\cite{Wahrburg_ECC2019} uses NNs to detect also intentional contacts  but limited to the upper and lower parts of the robot. 

This paper aims to devise an overall architecture  where the behavior of the robot is adjusted according to the type of interaction between human and robot and the robot task.
The robot is endowed with the ability of autonomously performing tasks, while the human operator is allowed to work side-by-side with the robot to perform additional tasks or to change robot configuration by physically interacting with it. Therefore, similarly to the papers cited above and as depicted in~\Cref{fig:scene}, both accidental and intentional contacts might arise and need to be detected and classified; this is achieved adopting an RNN-based approach that only exploits joint torque sensors and a localization system. Based on the output of the RNN-based modules, an avoidance strategy is undertaken in the case of accidental contact, while an admittance behavior is enforced to the robot in case of intentional contact.\\
With respect to existing approaches, the proposed solution advances the state-of-the-art in the following aspects: \emph{(i)}~the human-robot contact is detected and classified even in the case the robot is in contact with the environment; this is achieved without additional sensors and by resorting to GMMs for modeling the contact required by the robot task;
\emph{(ii)} a comprehensive control strategy is devised to handle both type of contacts while taking into account the contact point along the robot structure, the current task and the robot constraints.
 The paper builds on our earlier work~\cite{roman} introducing the following main differences: 
    \emph{(i)} the assumption about the knowledge of the contact point along the robot structure has been removed thanks to the introduction of a perception component;
\emph{(ii)} the reaction module has been extended allowing the human operator also to reconfigure the robot internal configuration; 
  \emph{(iii)} the detection and classification modules have been validated considering different users not involved in the training of the NNs;
   \emph{(iv)} the approach has been entirely validated on a real-world setup composed of a Kinova Jaco$2$ performing different tasks, and a vision system.

 \section{Preliminaries and problem formulation}

 \subsection{Robot model}
The following dynamical model is assumed for the robot
 \begin{equation}\label{eq:manipmodeljoint}
 \bfM(\bfq)\ddot\bfq+\bfc(\bfq,\dot\bfq)+\bfg(\bfq)\!=\!\bftau\!+\bftau_{\tasym}+\bftau_{h}
 \end{equation}
 where $\bfq\in\Re^{n} $ is the joint position vector, $\bftau\in\Re^{n}$ is the joint torque vector; $\bfM(\bfq)\in\Re^{n\times n}$
 is the symmetric positive definite inertia matrix, $\bfc(\bfq,\dot\bfq)\in\Re^{n\times n}$ is
 the centrifugal and Coriolis terms vector, $\bfg(\bfq)\in\Re^{n}$ is the gravity terms vector, \mbox{$\bftau_{\tasym}=\bfJ(\bfq)\t\bfh_{\tasym}$} is the  torque vector induced by the interaction  $\bfh_{\tasym}\in\Re^6$ with the environment to carry out a given task $\tasym$, with $\bfJ(\bfq)\in\Re^{6\times n} $  the Jacobian matrix at the end effector, $\bftau_{h} =  \bfJ_{\p}(\bfq)\t\bfh_{h}$  is the torque vector induced by the human wrench $\bfh_{h}\in\Re^6$ exerted at a generic point $P$ along the robot structure,
with \mbox{$\bfJ_{\p}(\bfq)\in\Re^{6\times n}$} the robot Jacobian matrix at this point.

We assume that the robot is able to  track a joint space reference trajectory $\bfq_{r}(t)\in\Re^{n} $, i.e. $\bfq_{r}\approx\bfq$. Note that such an assumption is commonly verified with off-the-shelf robots, for which built-in low-level controllers are integrated that generate the torque input $\tau$ in \cref{eq:manipmodeljoint}. In addition, in the case of custom platforms, standard linearizing control laws~\cite{hanbook_chapt8} can be implemented to the purpose. 
The robot model, thus, becomes
\begin{equation}\label{eq:mandyn}
\ddot \bfq = \bfu
\end{equation}
where  $\bfu_{}\in\Re^{n}$ is a virtual control input to be designed.

Finally, by virtue of \cref{{eq:mandyn}}, the well-known second order kinematic relationship at a generic point~$P$ on the robot structure gives
\begin{equation}\label{eq:kinemodel}
\ddot\bfx_\p=\bfJ_\p(\bfq)\ddot\bfq+\dispdot{\bfJ}_\p(\bfq, \dot\bfq)\dot\bfq=\bfJ_\p(\bfq)\bfu+\dispdot{\bfJ}_\p(\bfq, \dot\bfq)\dot\bfq
\end{equation}
where  $\bfx_\p=\mymatrix{\bfp_\p\t\,\, \bfvarphi_\p\t}\t\in\Re^{6}$ is the configuration of a frame centered in  $P$  with position $\bfp_\p$ and orientation $\bfvarphi_\p$. In the following, we omit subscript $\p$ when the kinematics is referred to the robot end effector. 
\subsection{Human torque and wrench estimation}
In order to detect and recognize the type of contact, the estimation of the human torque and wrench is needed. To this aim, ad-hoc hardware devices  can be integrated, e.g. \cite{Albini_IJRR2020}, or momentum-based observers that  only rely on on-board current/torque sensors can be leveraged~\cite{DeLuca_ICRA2005}. For the sake of generality, the latter approach is pursued in this work. 

Let us introduce the residual vector $\bfr(t)\in\Re^6$ defined as
\begin{equation}\label{eq:res}
\bfr(t)=\bfK\left(\int_{0}^{t}{(\bfalpha-\bftau-\bfr)d\tau}+\bfm(t)\right)
\end{equation}
with $\bfK\in\Re^{n\times n}$ a constant diagonal positive definite matrix,  $\bfm(t)=\bfM(\bfq)\dot\bfq$ the generalized momentum of the manipulator and 
$\bfalpha=\displaystyle \bfg-{1}/{2}\,\dot\bfq\t({\partial\bfM}/{\partial\bfq})\dot\bfq$.
By following the steps in~\cite{DeLuca_ICRA2005} and for 
high values of $\bfK$, the following holds
\begin{equation}\label{eq:r_appr}
\bfr(t)\approx(\bftau_{\tasym}+\bftau_{h}).
\end{equation}
Let us assume that an estimate $\hat \bftau_\tasym$ of $ \bftau_\tasym$  is available (more details in \Cref{sec:approx}) and that, in the case of human interaction, the contact point $P$ along the robot structure is estimated (more details in \Cref{sec:local}).
From \eqref{eq:r_appr}, an estimate $\hat{\bftau}_h$ 
of $\bftau_h$ is obtained as
\begin{equation}\label{eq:tauhest}
\hat{\bftau}_h(t) = \bfr(t)-\hat \bftau_\tasym 
\end{equation}
based on which { the components of ${\bfh}_h(t)$ not belonging to the null space of $\bfJ_\p\t$  can be retrieved}~\cite{BioRob_2012} as 
\begin{equation}\label{eq:hest}
    \hat{\bfh}_h(t) =  \left(\bfJ_\p \left(\bfq(t)\right)\t\right)^\dag \,\hat{\bftau}_h(t).
\end{equation}

 \subsection{Problem formulation and solution overview}\label{ssec:problem}
 Formally,  the following problem is addressed in the paper. 
 \begin{problem}\label{pr}
Consider the robot dynamics in~\eqref{eq:mandyn} and a task represented by a desired end effector trajectory  $\bfx_d(t)$ ($\dot\bfx_d(t),\ddot\bfx_d(t)$) which can possibly involve  interaction $\bfh\ta$ with the environment. Assume that a human operator 
can intentionally or accidentally physically interact with any point of the robot structure, i.e. $\|\bftau_h\|>0$.
 The aim is to design a strategy enabling the robot to detect and classify possible contacts with the human and to undertake proper reaction behaviors, while complying with possible robot constraints. 
 \end{problem}
	  \begin{psfrags}
		\def\scal{0.8}  
		\def\scals{0.7} 
		\psfrag{b1}[cc][][\scal]{\begin{tabular}{@{}c@{}}Contact \\[-4pt] Detection \end{tabular} }
		\psfrag{b2}[cc][][\scal]{\shortstack[c]{Contact Type \\[-4pt]  Recognition}}
		\psfrag{b3}[cc][][\scal]{ Robot behavior} 
		\psfrag{gmm}[cc][][\scals]{\shortstack[c]{Nominal task \\ interaction approx. }}
		\psfrag{det}[cc][][\scals]{\shortstack[c]{Detection NN}}
		\psfrag{loc}[cc][][\scals]{\shortstack[c]{Contact point \\ localization}}
		\psfrag{rec}[cc][][\scals]{\shortstack[c]{Recognition NN}}
		\psfrag{ea}[cc][][\scals]{\shortstack[c]{Avoidance \\ behavior}}
		\psfrag{adm}[cc][][\scals]{\shortstack[c]{Admittance control \\ behavior}}
		\psfrag{aut}[cc][][\scals]{\shortstack[c]{Task execution \\ behavior}}
		\psfrag{o1}[cc][][\scals]{}
		\psfrag{o2}[cc][][\scals]{}
		\mytoppsfrag{8.0}{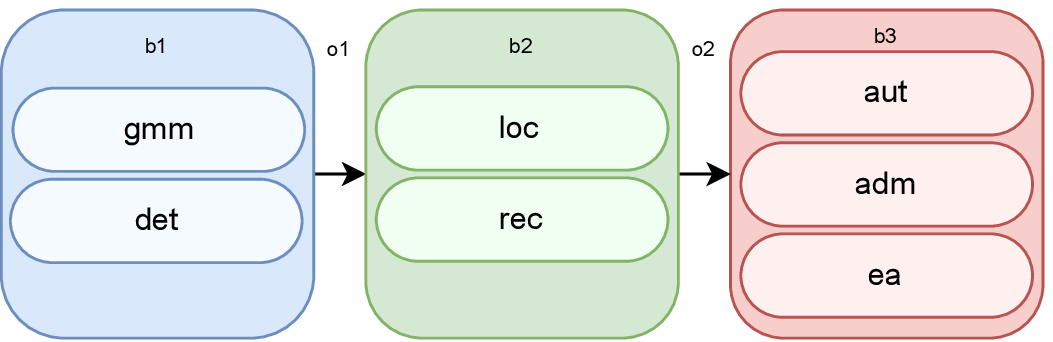}{-12pt}{Overview of the proposed approach to detect human contact  (in blue), recognize its type (in green) and react accordingly (in red).  }{fig:scheme} 
		\vspace{10pt}
	\end{psfrags}
	As shown in \Cref{fig:scheme}, a modular architecture composed of three blocks is proposed to solve the problem above. 
In detail,  a \textit{detection module} is in charge of assessing whether or not a contact with a human operator is taking place by considering that the robot might be interacting with the environment. Next, a \textit{recognition module} establishes the nature of the human interaction when contact is detected. To this aim, the contact point along the robot structure is localized, and the estimated wrench exerted by the human is evaluated according to~\eqref{eq:hest} to evaluate whether or not the contact is intentional. 
Finally, a \textit{behavior module} determines the robot virtual input $\bfu$ in~\eqref{eq:mandyn}.
In detail, the robot is endowed with three basic behaviors, namely  task execution, admittance control and avoidance behaviors, that are selected considering  the type of contact (if any) and the current task. 
\section{Detection and Recognition Modules}\label{sec:blueblockgreen}
This section details the components related to the detection and recognition modules in  \Cref{fig:scheme}. 
\subsection{Nominal task interaction approximation}\label{sec:approx}
 In order to assess whether or not a contact with a human operator is occurring, we define an approximation $\hat\bftau\ta$ (blue block of \Cref{fig:scheme}) of the expected external torque $\bftau\ta$, arising from the  interaction with the environment for task $\tasym$. Indeed, based on this estimation and on the   estimated overall external torque in~\eqref{eq:r_appr},  the estimated torques induced by the \textit{human} interaction $\hat\bftau_h$ can be derived.

We propose to use Gaussian Mixture Models to approximate the nominal wrench profile. In this way, a \textit{rough} definition of the expected nominal profile can be derived on the basis of   \textit{few} demonstrations of the nominal interaction with the environment, as shown in the experiments in \Cref{sec:exp}. 

Let us introduce the extended vector of the nominal wrench of task~$\tasym$ at time $t$ as $\hn\sta{t} = [t\,\,\, \bfh\sta{n}(t)\t]\t\in\Re^7$. This can be probabilistically  modeled as a mixture of $K\ta$ Gaussian distributions as follows
    \begin{equation}\label{eq:gmm}
        p(\hn\sta{t}) = \displaystyle \sum \nolimits_{k=1}^{K\ta} \pi\sta{k} \,\mathcal{N} (\hn\sta{t} | \bfmu\sta{k}, \bfSigma\sta{k})
    \end{equation}
where $\mathcal{N} (\hn\sta{t} | \bfmu\sta{k}, \bfSigma\sta{k})$ is the Gaussian distribution with mean $\bfmu\sta{k}\in\Re^{7}$ and covariance $\bfSigma\sta{k}\in\Re^{7\times 7}$ and $\pi\sta{k}$ is the $k$\ts th mixture coefficient such that $\pi\sta{k} \in[0,1]$ and $\sum_{k=1}^{K\ta}\pi\sta{k} = 1$.

The  Expectation-Maximization (EM) algorithm~\cite{Redner_EM} is then leveraged to estimate the GMM parameters $\bfmu\sta{k}$, $\bfSigma\sta{k}$  and $\pi\sta{k}$ $\forall k\in\{1,\ldots, K\ta\}$ based on demonstrations of $\hn\sta{t} \,\forall t$. To gather the latter, the task $\tasym$ is executed few times in the absence of human wrench, and the residual vector $\bfr(t)$ in~\eqref{eq:res} is recorded; the latter coincides with the estimate of $\bftau_{\tasym}(t)$ when $\bftau_h =\zero_n $, being $\zero_n$ the null vector of $n$ elements.  Each datapoint $\bfh\sta{n}(t)$ in the demonstrations of $\hn\sta{t}$ is then  obtained as 
\begin{equation}
\bfh\sta{n}(t) = \left(\bfJ \left(\bfq(t)\right){\t}\right)^\dag  \,\bfr(t).
\end{equation}
The choice of approximating the expected wrench $\bfh\sta{n}$  instead of the expected external torque $\bftau_{\tasym}$ enables  the proposed  methodology  to be independent from the joint space configuration while executing the task. 

Finally, Gaussian Mixture Regression, as done for instance in~\cite{Calinon_IROS20013}, is  used to get the probabilistic model of $\bfh\sta{n}(t)$ at each time $t$, with mean  $\hat{\bfh}_{\tasym,n}(t) = \mathbb{E}( {\bfh}_{\tasym,n}(t))$. The estimate $\hat \bftau_\tasym(t)$ in \cref{eq:tauhest}  is then achieved, i.e. \mbox{$\hat \bftau_\tasym(t) = \bfJ(\bfq(t))\t\,\hat{\bfh}_{\tasym,n}(t)$}.
\subsection{Detection and recognition classifiers}\label{ssec:classbluegreen}
As mentioned above, the estimated torques $\hat \bftau_\tasym$ are exploited to compute  the human torques $\hat\bftau_h$. Based on the latter, two RNNs are trained to determine the presence of a human contact and its type. 
 In detail, the first network, referred to as \textit{detection NN}, outputs the information on whether or not human contact is occurring, while the second network, referred to as \textit{recognition NN}, outputs the information on whether the contact is intentional or accidental.  In light of the binary classification problems, the two networks are trained to minimize the log loss. The same architecture is devised for both the networks and it comprises: \emph{(i)} a recurrent  Long Short Term Memory layer, that enables to learn long-term dependencies between data samples in the time series, \emph{(ii)}  a fully connect layer, \emph{(iii)} a logistic activation layer that outputs the prediction. 

The training dataset of the detection NN consists of labeled time series representing the norm of the estimated human torque, i.e. $\|\hat{\bftau}_h(t)\|$ in \cref{eq:tauhest}, while estimated human wrenches are considered for the recognition NN, i.e. the dataset consists of the datapoints $\|\hat{\bfh}_h(t)\|$ in  \cref{eq:hest}. The use of norm values allows to  take into account the intensity and variation of the human interaction ignoring the directions in which the interaction occurs. 
The datasets were collected by voluntarily and accidentally  interacting with different points along the robot structure during free space motion (i.e. with $\|\bftau\ta\|=0$) and recording the labeled respective quantities. 
Note that the datasets only contain data acquired with no environment interaction; however, the proposed approach is able to tackle the case of $\|\bftau\ta\|>0$  thanks to the GMM modeling in \Cref{sec:approx} and without wrist force/torque sensors. In this way, an arbitrary number of tasks can be performed that require interaction with the environment \textit{without} the need of re-training the networks and adapting their datasets. Finally, note that  the generalization capabilities of the networks allows to tackle, without manually tuned thresholds,  noisy measures, model uncertainties in~\eqref{eq:manipmodeljoint}  and task interaction estimation inaccuracies which generally leads to not accurate human torque estimation, i.e. it generally holds $||\hat\bftau_h||>0$ even if no human-robot contact is happening.

\subsection{Contact Point Localization}\label{sec:local}
We foresee the presence of a procedure  to localize the contact point~$P$, needed to estimate the human wrench in~\eqref{eq:hest} used by the recognition classifier and the behavior module.  
Not to limit the possible approaches and since we do not aim at improving localization techniques, we do not fix a specific  procedure but any appropriate solution can be applied to the purpose. As an example, and as considered in the experiments in \Cref{sec:exp}, an external vision algorithm can be integrated which tracks the human motion and determines the contact point. Alternatively, different ad-hoc sensors, such as  IMU sensors~\cite{Yuan_Sensors2014}, can be integrated to the purpose.

\section{Robot behavior module}\label{sec:robobehav}
Three basic robot behaviors (red block of \Cref{fig:scheme}) are envisaged which are activated depending on the type (if any) of interaction. In the following, we introduce the individual behaviors and we show their activation rationale, as well as how possible robot and task constraints can be accounted for.    
\subsection{Robot behaviors}
\subsubsection{Task execution behavior}
The task execution behavior allows the robot to perform its desired task which, as stated in Problem~\ref{pr}, is given by the desired trajectory at the effector $\bfx_d(t)$. 
The well-known closed loop inverse kinematic law is  leveraged to define the respective  virtual input $\bfu$ in \cref{eq:mandyn}: 
\begin{equation}\label{eq:ddqt:1}
\!\!\!\bfu_n = \ddot \bfq_{\tasym_n} +\ddot{\bfq}_{N},\quad \ddot \bfq_{\tasym_n}\!\!= \!\bfJ^\dag\!\!\left(\ddot \bfx_d\!+\!\bfK_{d}\dot{\tilde{\bfx}}+\!\bfK_{p} {\tilde{\bfx}} - \dispdot{\bfJ} \dot{\bfq} \right) 
\end{equation}
where ${\tilde {\bfx}}(t)=(\bfx_d(t)-\bfx(t))\in\Re^6$ is the task tracking error, $\bfK_{d}\in\Re^{6\times 6}$, $\bfK_{p}\in\Re^{6\times 6}$ are positive definite matrices, and $\ddot{\bfq}_{N}\in\Re^{n}$ is an arbitrary vector of joint accelerations which can be used for secondary tasks.

Differently from our previous work~\cite{roman}, we include the possibility for the human operator to reconfigure the \textit{internal} structure of the robot, i.e. without altering the robot task. To this aim, we define the following joint accelerations $\ddot{\bfq}_{N}$~\cite{deLuca_1992} in~\cref{eq:ddqt:1}, when an intentional interaction is detected in a  point $P$ different from the end effector position: 
\begin{equation}\label{eq:ddqn}
\resizebox{\linewidth}{!}{$
  \ddot \bfq_{N}= \left(\bfJ_\p(\bfI_n-\dispbar{\bfJ}^\dag\dispbar{\bfJ})\right)^\dag\!\biggl[\bfM_d^{-1}\!\left(-\bfD_{d}\dot{\bfx}_\p  + \hat\bfh_h\right)-\dispdot{\bfJ}_\p \dot{\bfq}-\bfJ_\p\ddot \bfq_{\tasym_n}\biggr] 
$}
\end{equation}
where $\bfM_d,\bfD_{d}\in\Re^{6\times 6}$ are positive definite matrix gains, and  $\dispbar{\bfJ}$ coincides with the positional Jacobian matrix at the end effector, if the task orientation can be relaxed, and with the full Jacobian matrix $\bfJ$  (in case of redundant robots), otherwise. 
The formulation in~\eqref{eq:ddqn} makes the robot compliant at the contact point, enabling the internal reconfiguration by the human, while preserving the respective components of the  desired task. 

\subsubsection{Admittance control behavior}\label{ssec:admittance}
The admittance control behavior allows the human to guide the robot motion according to a mass-damper model. In particular, when the human intentionally interacts with the robot end effector, we assume that he/she wants to intervene in the robot task and correct its execution. The following admittance strategy is thus used for the input $\bfu$ in \cref{eq:mandyn}: 
\begin{equation}\label{eq:ddqt:2}
\!\!\!\!\bfu_a = \ddot \bfq_{\tasym_a} +\ddot{\bfq}_{N},\quad\!\!\!\! \ddot \bfq_{\tasym_a}\!\!= \!\!\bfJ^\dag\!\left(\bfM_d^{-1}\!\left(-\bfD_{d}\dot{\bfx}  + \hat\bfh_h\right)\!-\!\dispdot{\bfJ} \dot{\bfq} \right) 
\end{equation}
where, as above, $\ddot{\bfq}_{N}\in\Re^{n}$ is an arbitrary vector of joint accelerations, while $\bfM_d,\bfD_{d}\in\Re^{6\times 6}$  are the positive definite desired inertia and damping, respectively. 
By replacing~\eqref{eq:ddqt:2} in \eqref{eq:kinemodel}, the closed loop model is derived: 
\begin{equation}\label{eq:cltasknom-adm}
\bfM_d\,\ddot{{\bfx}}+\bfD_d\,\dot{{\bfx}}=\hat{\bfh}_h.
\end{equation}

\subsubsection{Avoidance behavior}\label{ssec:avoidance}

As in our earlier work~\cite{roman}, we envisage that an avoidance behavior is activated when an accidental collision occurs. To this aim, the concept of safety field $F(t)$ in~\cite{Lippi_TCST2020} is leveraged which assesses the level  of human safety with respect to the robot in a comprehensive manner, i.e. the entire human body and robot structure are taken into account. 
The objective of the avoidance behavior is thus to recover a safety condition $F\geq F_d$ after the occurrence of a collision, with $F_d$ a positive threshold to be tuned~\cite{Lippi_TCST2020}. In order to achieve this objective, the following virtual input $\bfu_o$ is defined
\begin{equation}\label{eq:ddqt:3}
\bfu_o = \ddot \bfq_{\tasym_o} +\ddot{\bfq}_{N},\,\,\, \ddot\bfq_{\tasym_o} = \bfJ^\dag_F\!\left(k_d\dot \Delta F +k_p\Delta F\!-\!\dispdot{\bfJ}_F\dot\bfq\right) 
\end{equation}
where $\bfJ_F\in\Re^{1 \times n}$ is the safety field Jacobian matrix such that $\dot F = (\partial F/\partial\bfq)\dot\bfq=\bfJ_F\dot\bfq$, \mbox{$\Delta F = \min(0, F_d-F)$}, $k_{p},k_{d}$  are positive gains,  and $\ddot{\bfq}_{N}\in \Re^n$ is an additional acceleration vector. By replacing \eqref{eq:ddqt:3} in \eqref{eq:mandyn}, the safety field dynamics is
\begin{equation}\label{eq:cltasknom-F}
\ddot \Delta F + k_d\dot \Delta F +k_p\Delta F= 0
\end{equation}
implying the asymptotic convergence of $\Delta F$ to the origin.

\subsection{Behavior selection and robot constraints}
The Finite State Machine (FSM) in \Cref{fig:fsm} regulates the activation of the appropriate behavior. More specifically, the robot task is carried out according to \eqref{eq:ddqt:1}-\eqref{eq:ddqn} as long  as no human interaction is detected, or if a force is voluntarily applied along the robot structure to internally reconfigure it. Then, in the case an intentional contact is recognized at the end effector, the admittance behavior in \eqref{eq:ddqt:2} is selected which enables the human to adjust the robot task. The robot persists in this state until the human exerts wrenches at the contact point. Finally, if an accidental contact is recognized,  the robot selects the avoidance behavior to increase the human safety field $F$. The robot persists in this state as long as a safety condition is not restored, i.e. as long as $F<F_{min}$, with $F_{min}>F_d$ a scalar threshold. 
\begin{psfrags}
\setlength{\textfloatsep}{0.7\baselineskip plus 0.2\baselineskip minus 0.5\baselineskip}
		\def\scal{0.8}  
		\def\scals{0.7} 
		\psfrag{s1}[cc][][\scal]{\shortstack[c]{Task \\exec. (eqs. \\ \eqref{eq:ddqt:1}-\eqref{eq:ddqn})}}
		\psfrag{s3}[cc][][\scal]{\shortstack[c]{Avoidance\\(\cref{eq:ddqt:3})}}
		\psfrag{c3}[cc][][\scals]{Accidental contact}
		\psfrag{c4}[cc][][\scals]{$F \geq F_{min}$}
		\psfrag{s2}[cc][][\scal]{\shortstack[c]{Admittance\\(\cref{eq:ddqt:2})}}
		\psfrag{c1}[cc][][\scals]{\shortstack[c]{Intentional contact at \\ end effector}}
		\psfrag{c2}[cc][][\scals]{No contact}
\mypsfrag{5.5}{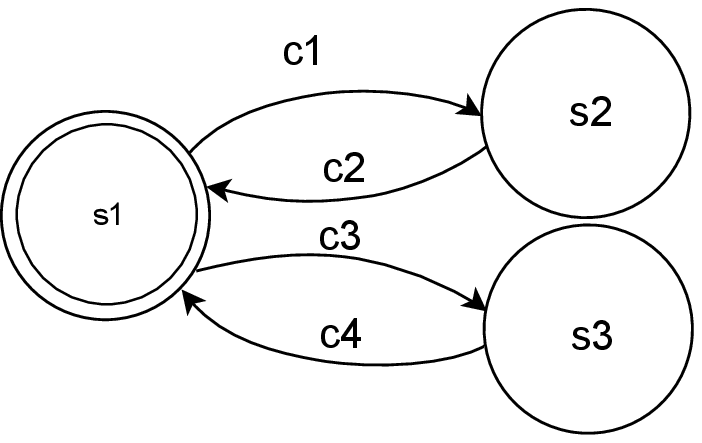}{-17pt}{FSM for robot's behaviors activation. }{fig:fsm} 
	\end{psfrags}

In order to handle possible constraints, depending on the task itself and on the nature of the interaction (see  \Cref{sec:tasks-exp} for examples), 
 the system in~\eqref{eq:mandyn} is rewritten in the form 
\begin{equation}
\begin{aligned}
\dot\bfxi_q&=\bff(\bfxi_q) +\bfg(\bfxi_q)(\bfu)=\mymatrix{\Zero_n & \bfI_n\\ \Zero_n & \Zero_n}\bfxi_q+\mymatrix{\Zero_n \\ \bfI_n}\bfu    
\end{aligned}
\end{equation}
 with $\bfxi_q = [\bfq\t \,\,\,\dot\bfq\t]\t\in\Re^{2n}$, and $\Zero_m$ ($\bfI_m$) the $m\times m$ null (identity) matrix. 
We express the  $i$\ts th constraint as
\begin{equation}\label{eq:constr}
 \bfphi_i(\bfxi_x(\bfxi_q))\ge 0
\end{equation}
where $\bfxi_x = [\bfx_\p\t \,\,\,\dot\bfx_\p\t]\t\in\Re^{12}$  and $\bfphi_i(\cdot)$ is a continuous scalar function. In order to have these constraints satisfied, the CBF approach~\cite{Ames_CDC} is adopted.
Based on~\cite{Ames_CDC}, the control input $\bfu^\star$ which achieves the task function subject to  constraints can be computed as the solution of the following Quadratic Program problem
\begin{equation}\label{eq:optpr}
\begin{aligned}
\bfu^\star&=\displaystyle \argmin_\bfu  \frac{1}{2}\left(\bfu-\bfu_{(\cdot)}\right)\t\bfQ\left(\bfu-\bfu_{(\cdot)}\right)\\
s.t. &\quad L_f \phi_i +L_g \phi_i\bfu \geq -\gamma(\phi_i(\bfxi)),\,\,  \forall i 
\end{aligned}
\end{equation}
where $\bfu_{(\cdot)}$ is the desired input computed according to eqs.~\eqref{eq:ddqt:1}, \eqref{eq:ddqt:2} or \eqref{eq:ddqt:3} in dependence of the FSM state, 
 $\bfQ\in\Re^{n\times n}$ is a positive definite matrix, $\gamma(\cdot)$ is  an extended $\mathcal K_\infty$ class function and $L_f \phi_i$, $L_g \phi_i$ are the Lie derivatives of $\phi_i$ with respect to $f$ and $g$, respectively.

\section{Experiments}\label{sec:exp}
 The experimental validation of the approach is presented in this section. The entire experiment execution is shown in the accompanying video.

\subsection{System architecture and robot's tasks}\label{sec:tasks-exp}
The system setup, illustrated  in \Cref{fig:scene}, is composed of a Kinova Jaco$2$ robotic arm with $7$~DOFs ($n=7$) and a Microsoft Kinect One RGB-D sensor. All the algorithms related to the contact classification and to the control run on \Matlab~$2020$ and exchange information with Robot Operating System (ROS). A vision-based localization component (green block in \Cref{sec:local}) is also included in ROS. 
In detail, a human skeleton tracking  algorithm\footnote{\url{https://github.com/mcgi5sr2/kinect2_tracker}} is adopted which, combined with the robot configuration information, allows to identify the point $P$ along the robot structure where human contact occurs.

Concerning the robot arm, the following collaborative tasks are considered.

\subsubsection{Object pick and place} The robot can manipulate bottles and mugs. 
To prevent the liquid from spilling out of the objects, we consider the following orientation constraints:
\begin{equation}\label{eq:constr-or}
    \begin{aligned}
    \phi_{o,i}^l &=  \varphi_i- (\varphi_{d,i} -\Delta\varphi_i) \geq 0 \\ 
    \phi_{o,i}^u &=  (\varphi_{d,i} + \Delta\varphi_i)-\varphi_i \geq 0  
    \end{aligned}
    \quad i = 1,2,3
\end{equation}
which are in the same form as in \eqref{eq:constr}, being $\bfvarphi_{d}\in\Re^3$ a desired orientation, $\Delta\varphi_i$  a positive tolerance and $(\cdot)_i$ the $i$\ts th component of the respective vector. The above constraints have a relative degree equal to $2$ which are handled as in~\cite{basso2020taskpriority}.

\subsubsection{Pouring task} This task foresees the robot holding a mug and the human filling it. The nominal wrench profile due to the filling is approximated with $K_\tasym = 5$ Gaussians and by using $5$ randomly selected demonstrations. The GMM parameters for the EM algorithm are initialized by resorting to a K-means clustering algorithm. 
Also in this case, the end effector orientation is constrained as in~\eqref{eq:constr-or}.
\subsubsection{Table cleaning} The robot executes a periodical motion on the table surface holding a cylindrical sponge (see \Cref{fig:scene}). 
The force exerted on the table is approximated through GMMs by using $5$ randomly selected demonstrations with, as before, $K_\tasym = 5$ and K-means algorithm for initialization.

Further constraints in the CBF framework are introduced in all the tasks to limit the robot workspace and avoid collisions with the table, i.e. $\phi_t = p_z - h_t\ge0$ in \eqref{eq:constr} being $h_t$ the height of the table. 
Note that the human is always allowed to physically interact with the robot end effector, changing its configuration, or with the robot structure, changing its internal joint configuration.  
In this way, the user can help the robot to accomplish the task or to avoid possible collisions with obstacles not detected by the vision system.

\subsection{Classification results}
\begin{table}[]
\resizebox{\linewidth}{!}{%
\begin{tabular}{cccccccc}
                          & \multicolumn{7}{c}{Training user}
                          \\
                          & \multicolumn{3}{c}{Detection NN}                                                                                                                                                             & \multicolumn{1}{l}{}      & \multicolumn{3}{c}{Recognition NN}                                                                                                                                                           \\
                          & $nc$                                                  & $wc$                                                  &                                                                              &                           & $ic$                                                  & $ac$                                                  &                                                                              \\ \hhline{~---~---} 
                          
\multicolumn{1}{c|}{$nc$} & \multicolumn{1}{c|}{\cellcolor[HTML]{B1DDF0}$4697$}   & \multicolumn{1}{c|}{$373$}                            & \multicolumn{1}{c|}{\cellcolor[HTML]{BAC8D3}$92.6\%$}                        & \multicolumn{1}{c|}{$ic$} & \multicolumn{1}{c|}{\cellcolor[HTML]{B1DDF0}$3979$}   & \multicolumn{1}{c|}{$135$}                            & \multicolumn{1}{c|}{\cellcolor[HTML]{BAC8D3}$96.7\%$}                        \\ \hhline{~---~---} 
\multicolumn{1}{c|}{$wc$} & \multicolumn{1}{c|}{$139$}                            & \multicolumn{1}{c|}{\cellcolor[HTML]{B1DDF0}$2930$}   & \multicolumn{1}{c|}{\cellcolor[HTML]{BAC8D3}$95.5\%$}                        & \multicolumn{1}{c|}{$ac$} & \multicolumn{1}{c|}{$555$}                            & \multicolumn{1}{c|}{\cellcolor[HTML]{B1DDF0}$1826$}   & \multicolumn{1}{c|}{\cellcolor[HTML]{BAC8D3}$76.7\%$}                        \\ \hhline{~---~---} 
\multicolumn{1}{c|}{}     & \multicolumn{1}{c|}{\cellcolor[HTML]{BAC8D3}$97.1\%$} & \multicolumn{1}{c|}{\cellcolor[HTML]{BAC8D3}$88.7\%$} & \multicolumn{1}{c|}{\cellcolor[HTML]{69838F}{\color[HTML]{FFFFFF} $93.7\%$}} & \multicolumn{1}{c|}{}     & \multicolumn{1}{c|}{\cellcolor[HTML]{BAC8D3}$87.8\%$} & \multicolumn{1}{c|}{\cellcolor[HTML]{BAC8D3}$93.1\%$} & \multicolumn{1}{c|}{\cellcolor[HTML]{69838F}{\color[HTML]{FFFFFF} $89.3\%$}} \\ \hhline{~---~---} 
\multicolumn{1}{l}{}      & \multicolumn{1}{l}{}                                  & \multicolumn{1}{l}{}                                  & \multicolumn{1}{l}{}                                                         & \multicolumn{1}{l}{}      & \multicolumn{1}{l}{}                                  & \multicolumn{1}{l}{}                                  & \multicolumn{1}{l}{}                                                         \\
                          & \multicolumn{7}{c}{Novel users} 
                          \\
                          & $nc$                                                  & $wc$                                                  &                                                                              &                           & $ic$                                                  & $ac$                                                  &                                                                              \\ \hhline{~---~---} 
\multicolumn{1}{c|}{$nc$} & \multicolumn{1}{c|}{\cellcolor[HTML]{B1DDF0}$6120$}   & \multicolumn{1}{c|}{$456$}                            & \multicolumn{1}{c|}{\cellcolor[HTML]{BAC8D3}$93.1\%$}                        & \multicolumn{1}{c|}{$ic$} & \multicolumn{1}{c|}{\cellcolor[HTML]{B1DDF0}$21041$}  & \multicolumn{1}{c|}{$2647$}                           & \multicolumn{1}{c|}{\cellcolor[HTML]{BAC8D3}$88.8\%$}                        \\ \hhline{~---~---} 
\multicolumn{1}{c|}{$wc$} & \multicolumn{1}{c|}{$2836$}                           & \multicolumn{1}{c|}{\cellcolor[HTML]{B1DDF0}$19745$}  & \multicolumn{1}{c|}{\cellcolor[HTML]{BAC8D3}$87.4\%$}                        & \multicolumn{1}{c|}{$ac$} & \multicolumn{1}{c|}{$3669$}                           & \multicolumn{1}{c|}{\cellcolor[HTML]{B1DDF0}$7016$}   & \multicolumn{1}{c|}{\cellcolor[HTML]{BAC8D3}{\color[HTML]{333333} $65.7\%$}} \\ \hhline{~---~---} 
\multicolumn{1}{c|}{}     & \multicolumn{1}{c|}{\cellcolor[HTML]{BAC8D3}$68.3\%$} & \multicolumn{1}{c|}{\cellcolor[HTML]{BAC8D3}$97.7\%$} & \multicolumn{1}{c|}{\cellcolor[HTML]{69838F}{\color[HTML]{FFFFFF} $88.7\%$}} & \multicolumn{1}{c|}{}     & \multicolumn{1}{c|}{\cellcolor[HTML]{BAC8D3}$81.2\%$} & \multicolumn{1}{c|}{\cellcolor[HTML]{BAC8D3}$72.6\%$} & \multicolumn{1}{c|}{\cellcolor[HTML]{69838F}{\color[HTML]{FFFFFF} $81.6\%$}} \\ \hhline{~---~---} 
\end{tabular}
}
\vspace{-2pt}
\caption{Confusion matrices of the RNNs with data acquired by the user involved in the training dataset (top part) and with different users not involved in the training dataset (bottom part). }

\label{tab:conf}
\end{table}
This section collects the classification results of the two RNNs, for which $100$ hidden units have been considered. 
In particular, the detection NN determines if a datapoint in the time series belongs to the class {\it no contact} ($nc$) or {\it with contact} ($wc$), while the recognition network establishes if, in the case of contact,  a datapoint belong to the class {\it intentional contact} ($ic$) or {\it accidental contact} ($ac$).   

The training dataset of the  detection (recognition) NN consists of  $43487$ ($15647$) time samples, representing  $869.7$~s ($312.9$~s) with time step $T = 0.02$~s,  equally distributed  in their respective two classes. 
Concerning the test set, we evaluated the accuracy both with the user that was involved in the collection of the training dataset, referred to as \textit{training user}, and with other four users \textit{not} involved in the training process, referred to as \textit{novel users}.  
Table~\ref{tab:conf} reports the confusion matrices of the classifiers for the training user (top part) and the novel users (bottom part). 
We observe that all the RNNs achieve good overall accuracy which is, for the training user,  $\approx 94\%$ for the detection NN and $\approx 90\%$ for the recognition NN. The lower accuracy of the latter is motivated by the intrinsic greater complexity of the problem of recognizing the type of contact rather than identifying \textit{any} possible contact. Remarkably, the results also confirm the generalization capabilities of the chosen NNs which allow classification also with novel users. In detail, an overall accuracy equal to $\approx90\%$ and $\approx82\%$   
is obtained for the detection and recognition classifications, respectively, implying a decrease of performance of only $\approx4\%$ and $\approx8\%$, compared to the training user. 
Finally, note that the results in \Cref{tab:conf} are obtained with a sample-by-sample evaluation;  however, a certain delay for detecting and classifying a contact always occurs as the networks obviously need some samples before being able to correctly classify. This motivates the classification inaccuracies in \Cref{tab:conf} but, as demonstrated in the following case studies, it does not  undermine the human-robot collaboration. 

\subsection{Case studies and experimental results}
We consider two case studies for the experimental validation which involve different tasks and human interactions. 
In both case studies, we used the following parameters: $\bfK=50\bfI_7$ in \cref{eq:res}, $\bfK_d=5\bfI_6$, $\bfK_p=6\bfI_6$ in~\eqref{eq:ddqt:1}, $\bfM_d=5\bfI_6$, $\bfD_d=100\bfI_6$ in~\eqref{eq:ddqn} and~\eqref{eq:ddqt:2}, $k_d=5$, $k_p=6$ in~\eqref{eq:ddqt:3}, $\bfQ=\bfI_7$ in~\eqref{eq:optpr}. Moreover, we set $F_d = 10$ in~\eqref{eq:ddqt:3} to ensure a minimum distance $\approx 0.4$~m (see~\cite{Lippi_TCST2020} for details) between every point of the robot and the human operator, and $F_{min} = 11$ in the FSM. 

\subsubsection{Case study $1$} In this case study, we let the robot execute three different tasks, while the human interacts with it. At the beginning, the robot takes the cylindrical sponge and starts a cleaning task, during which the human intentionally interacts with its end effector to change its configuration. Once the cleaning task is completed, the robot starts a bottle pick and place operation. During this task execution, the human voluntarily interacts along the robot structure to modify the joint configuration while preserving the end effector position. Finally, a pick operation of a mug and a pouring task are performed. During each task, the respective constraints and possible expected environment interaction defined in \Cref{sec:tasks-exp} are taken into account.  

Figure~\ref{fig:plot_case_1} summarizes the results of this case study. In detail, proceeding from the top to the bottom, it reports the robot end effector trajectory $\bfx$, the nominal interaction wrench $\hat\bfh\ta$, the norm of estimated human torque $\|\hat\bftau_h\|$ and the classification output of the RNNs (in blue) compared to the Ground Truth (GT, in green). 
Initially, the robot autonomously performs the cleaning task (highlighted with red boxes in the plots) and executes the desired periodic motion (first plot) which generates interaction wrenches with the table. The use of the nominal wrench profile (second plot) modeled via GMM allows to estimate the human torque (third plot) and to properly recognize that, in the initial phase, only interaction with the environment is occurring while 
no human contact is present (last plot). Once the human intentionally interacts with the end effector at $t\approx 21$~s, it is recognized by the RNNs  (with a delay  $<1$~s) and the admittance behavior is activated. This makes the robot end effector compliant towards the human wrench and its trajectory is modified as shown in the first plot. 
 At the end of the interaction, the execution of the desired tasks is restored and the pick and place operations are carried out (highlighted with green boxes) from $t\approx 27$~s to $t\approx 100$~s.  In this phase, an intentional  contact on link $4$ of the robot is recognized via the localization component at $t\approx 70$~s, which leads the robot to change its joint configuration according to~\eqref{eq:ddqt:1} and~\eqref{eq:ddqn} (see accompanying video) while preserving its end effector position, as shown in the top plot. Orientation variables are also limited according to~\eqref{eq:constr-or} with  $\bfvarphi_d = [-1.4,\,1.6,\,-2.9]\t$ and $\Delta\varphi_i = 0.2\,\forall i$. 
Finally, the pouring task (blue box in the plots) is executed starting from $t\approx 100$~s during which the respective GMM model is exploited and no contact with the human is detected. 

\begin{psfrags}
	\def\scal{0.7}  
	\def\scalnum{0.55}
	\def\scalleg{0.5}
    \input{numpsfrags}

	\psfrag{tau}[cc][][\scalleg]{ $\|\hat\bftau_h\|$}
	\psfrag{det}[cc][][\scalleg]{ RNN}
	\psfrag{gt}[cc][][\scalleg]{ GT}
	\psfrag{x1}[cc][][\scalleg]{$p_1$}
	\psfrag{x2}[cc][][\scalleg]{$p_2$}
	\psfrag{x3}[cc][][\scalleg]{$p_3$}
	\psfrag{x4}[cc][][\scalleg]{$\varphi_1$}
	\psfrag{x5}[cc][][\scalleg]{$\varphi_2$}
	\psfrag{x6}[cc][][\scalleg]{$\varphi_3$}
	\psfrag{f1}[cc][][\scalleg]{$\hat{h}_{\tasym,1}$}
	\psfrag{f2}[cc][][\scalleg]{$\hat{h}_{\tasym,2}$}
	\psfrag{f3}[cc][][\scalleg]{$\hat{h}_{\tasym,3}$}
	\psfrag{f4}[cc][][\scalleg]{$\hat{h}_{\tasym,4}$}
	\psfrag{f5}[cc][][\scalleg]{$\hat{h}_{\tasym,5}$}
	\psfrag{f6}[cc][][\scalleg]{$\hat{h}_{\tasym,6}$}
	\psfrag{c1}[cc][][\scalnum]{$nc$}
	\psfrag{c2}[cc][][\scalnum]{$ic$}
	\psfrag{c3}[cc][][\scalnum]{$ac$}
	\psfrag{l7}[cc][][\scalnum]{Link $7$}
	\psfrag{l4}[cc][][\scalnum]{Link $4$}
	\psfrag{t}[cc][][\scal]{ $t[s]$}
    \mypsfrag{7.8}{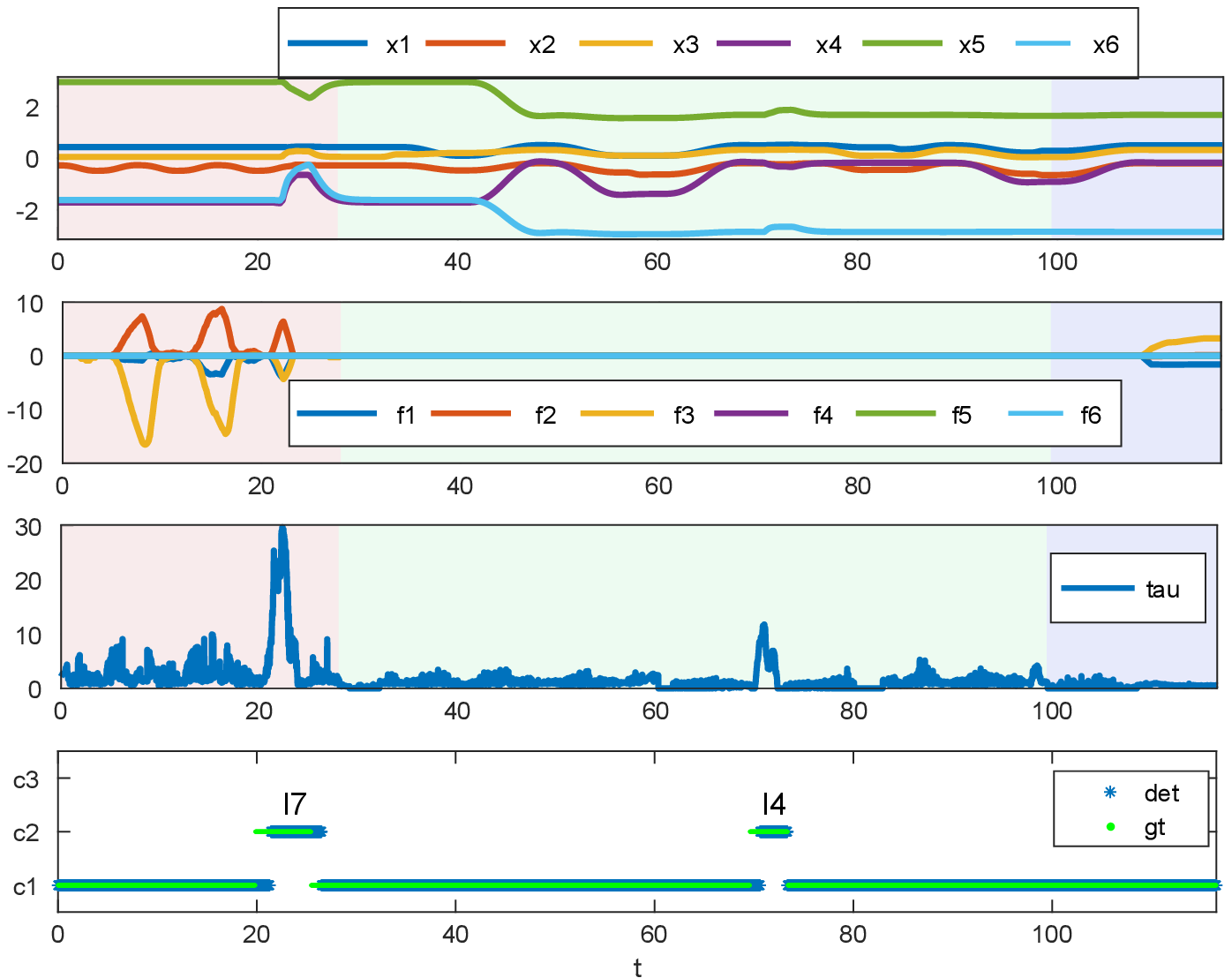}{-10pt}{Case study $1$. From the top: end effector trajectory, nominal task interaction wrench, norm of the estimated human torques and classification results compared to ground truth.  Cleaning, pick and place and pouring phases are denoted in red, green and blue. }{fig:plot_case_1} 
\end{psfrags}

\subsubsection{Case study $2$} This case study aims to prove the effectiveness of the solution also with accidental contacts. It is structured as follows: the human first intentionally interacts with the robot end effector during a pick and place task, and then an accidental collision happens.

Figure~\ref{fig:plot_case_2} shows, from the top, the robot end effector trajectory $\bfx$, the norm of the estimated human torques $\|\hat\bftau_h\|$ and the classification results (in blue), compared to ground truth values (in green). Based on the estimated human wrench (second plot) and on the localization component, an intentional contact at the end effector  is recognized by the RNNs (third plot) at $t\approx 9$~s; then, the admittance behavior is activated, leading to a modification of the robot trajectory. The robot task execution is then restored when the interaction terminates and an accidental contact is detected on link~$7$ at $t\approx 33$~s. This activates the avoidance behavior which drives the robot away from the human operator in order to increase the safety (see also the accompanying video). 

\begin{psfrags}
	\def\scal{0.7}  
	\def\scalnum{0.55}
	\def\scalleg{0.5}
    \input{numpsfrags}
	\psfrag{tau}[cc][][\scalleg]{ $\|\hat\bftau_h\|$}
	\psfrag{det}[cc][][\scalleg]{ RNN}
	\psfrag{gt}[cc][][\scalleg]{ GT}
	\psfrag{x1}[cc][][\scalleg]{$p_1$}
	\psfrag{x2}[cc][][\scalleg]{$p_2$}
	\psfrag{x3}[cc][][\scalleg]{$p_3$}
	\psfrag{x4}[cc][][\scalleg]{$\varphi_1$}
	\psfrag{x5}[cc][][\scalleg]{$\varphi_2$}
	\psfrag{x6}[cc][][\scalleg]{$\varphi_3$}
	\psfrag{f1}[cc][][\scalleg]{$\hat{h}_{\tasym,1}$}
	\psfrag{f2}[cc][][\scalleg]{$\hat{h}_{\tasym,2}$}
	\psfrag{f3}[cc][][\scalleg]{$\hat{h}_{\tasym,3}$}
	\psfrag{f4}[cc][][\scalleg]{$\hat{h}_{\tasym,4}$}
	\psfrag{f5}[cc][][\scalleg]{$\hat{h}_{\tasym,5}$}
	\psfrag{f6}[cc][][\scalleg]{$\hat{h}_{\tasym,6}$}
	\psfrag{c1}[cc][][\scalnum]{$nc$}
	\psfrag{c2}[cc][][\scalnum]{$ic$}
	\psfrag{c3}[cc][][\scalnum]{$ac$}
	\psfrag{l7}[cc][][\scalnum]{Link $7$}
	\psfrag{t}[cc][][\scal]{ $t[s]$}
    \mypsfrag{7.8}{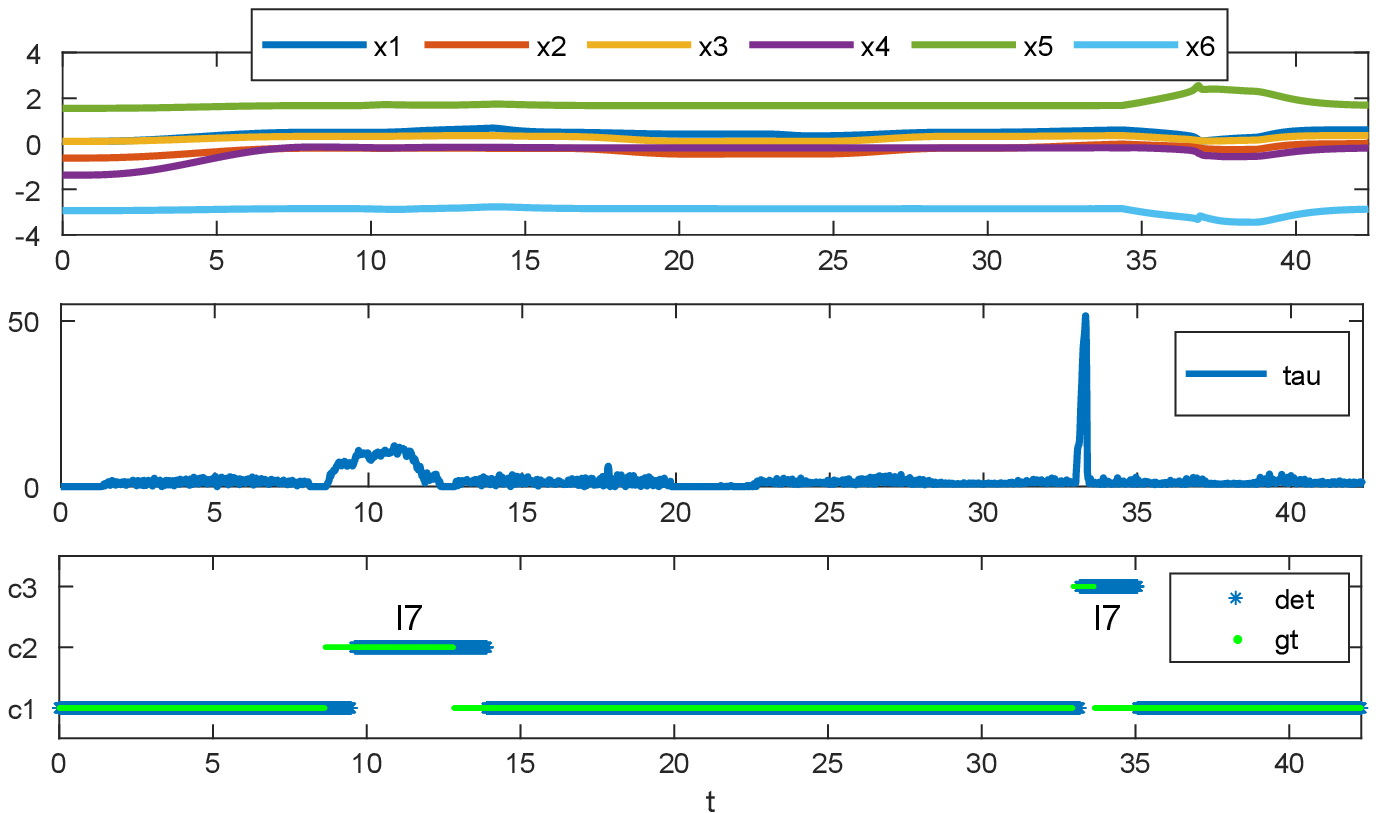}{-15pt}{Case study $2$. From the top: end effector trajectory, estimated human torques and classification results compared to ground truth.}{fig:plot_case_2} 
\end{psfrags}

\section{Conclusions}
We proposed a framework to detect human contact and discern if it is intentional or accidental, even when the robot's task requires contact with the environment by leveraging  RNNs for time series classification and GMMs for modeling the expected interaction wrench. Three basic robot behaviors are identified and activated on the basis of the human interaction. Experiments on a real platform corroborated the devised approach. As future work, we plan to include proactive actions depending on human and scene awareness and to consider multi-arm setups.

 \bibliography{biblio}
\end{document}

%% file: bmit2e.tex
\font\bfmath=cmmib10
\mathchardef\Gamma="7100
\mathchardef\Delta="7101
\mathchardef\Theta="7102
\mathchardef\Lambda="7103
\mathchardef\Xi="7104
\mathchardef\Pi="7105
\mathchardef\Sigma="7106
\mathchardef\Upsilon="7107
\mathchardef\Phi="7108
\mathchardef\Psi="7109
\mathchardef\Omega="710A

\mathchardef\alpha="710B
\mathchardef\beta="710C
\mathchardef\gamma="710D
\mathchardef\delta="710E
\mathchardef\epsilon="710F
\mathchardef\zeta="7110
\mathchardef\eta="7111
\mathchardef\theta="7112
\mathchardef\iota="7113
\mathchardef\kappa="7114
\mathchardef\lambda="7115
\mathchardef\mu="7116
\mathchardef\nu="7117
\mathchardef\xi="7118
\mathchardef\pi="7119
\mathchardef\rho="711A
\mathchardef\sigma="711B
\mathchardef\tau="711C
\mathchardef\upsilon="711D
\mathchardef\phi="711E
\mathchardef\chi="711F
\mathchardef\psi="7120
\mathchardef\omega="7121
\mathchardef\epsilon="7122

\mathchardef\varepsilon="7122
\mathchardef\vartheta="7123
\mathchardef\varpi="7124
\mathchardef\varrho="7125
\mathchardef\varsigma="7126
\mathchardef\varphi="7127
\mathchardef\imath="717B
\mathchardef\jmath="717C

\def\zero{{\bf 0}}
\def\Zero{{\mbox{\boldmath $O$}}}

\def\bfc{{\mbox{\boldmath $c$}}}

\def\bff{{\mbox{\boldmath $f$}}}
\def\bfg{{\mbox{\boldmath $g$}}}
\def\bfh{{\mbox{\boldmath $h$}}}

\def\bfm{{\mbox{\boldmath $m$}}}

\def\bfp{{\mbox{\boldmath $p$}}}
\def\bfq{{\mbox{\boldmath $q$}}}
\def\bfr{{\mbox{\boldmath $r$}}}

\def\bfu{{\mbox{\boldmath $u$}}}

\def\bfx{{\mbox{\boldmath $x$}}}

\def\bfD{{\mbox{\boldmath $D$}}}

\def\bfI{{\mbox{\boldmath $I$}}}
\def\bfJ{{\mbox{\boldmath $J$}}}
\def\bfK{{\mbox{\boldmath $K$}}}

\def\bfM{{\mbox{\boldmath $M$}}}

\def\bfQ{{\mbox{\boldmath $Q$}}}

\def\bfalpha{{\mbox{\boldmath $\alpha$}}}

\def\bfzeta{{\mbox{\boldmath $\zeta$}}}

\def\bfmu{{\mbox{\boldmath $\mu$}}}

\def\bfxi{{\mbox{\boldmath $\xi$}}}

\def\bftau{{\mbox{\boldmath $\tau$}}}

\def\bfphi{{\mbox{\boldmath $\phi$}}}

\def\bfvarphi{{\mbox{\boldmath $\varphi$}}}

\def\bfSigma{{\mbox{\boldmath $\Sigma$}}}

%% file: mymacro.tex
\def\Matlab{{\sc Matlab}}

\def\smallbfW{{\raise1.5pt\hbox{\mbox{\boldmath $_W$}}}}
\def\t{^{\rm T}}
\let\ts=\thinspace

\def\Re{\mathbb{R}}

\def\mypsfrag#1#2#3#4#5{
        \begin{figure}[htp]
           \begin{center}
              {\leavevmode
                 {\includegraphics[width=#1truecm]{#2.eps}}
              }
           \end{center}
           \vspace{#3}
           \caption{#4}
           \label{#5}
        \end{figure}
}
\def\mytoppsfrag#1#2#3#4#5{
        \begin{figure}[t!]
           \begin{center}
              {\leavevmode
                 {\includegraphics[width=#1truecm]{#2.eps}}
              }
           \end{center}
           \vspace{#3}
           \caption{#4}
           \label{#5}
        \end{figure}
}

\def\my4psfrag#1#2#3#4#5#6#7#8{
        \begin{figure}[htp]
        \begin{center}
            \begin{tabular}[h]{c c}
              {\leavevmode{\includegraphics[width=#1truecm]{#2.eps}}}
              &
              {\leavevmode{\includegraphics[width=#1truecm]{#3.eps}}} \\
              {\leavevmode{\includegraphics[width=#1truecm]{#4.eps}}}
              &
              {\leavevmode{\includegraphics[width=#1truecm]{#5.eps}}}
         \end{tabular}
           \vspace{#6}
           \caption{#7}
           \label{#8}
        \end{center}
        \end{figure}
}

\def\mydouble4psfrag#1#2#3#4#5#6#7#8{
        \begin{figure*}[htp]
        \begin{center}
            \begin{tabular}[h]{c c}
              {\leavevmode{\includegraphics[width=#1truecm]{#2.eps}}}
              &
              {\leavevmode{\includegraphics[width=#1truecm]{#3.eps}}} \\
              {\leavevmode{\includegraphics[width=#1truecm]{#4.eps}}}
              &
              {\leavevmode{\includegraphics[width=#1truecm]{#5.eps}}}
         \end{tabular}
           \vspace{#6}
           \caption{#7}
           \label{#8}
        \end{center}
        \end{figure*}
}

\def\mymatrix#1{\left[\begin{matrix}#1\end{matrix}\right]}

%% file: MATHSYM.TEX
\def\zero{\hbox{\bf 0}}
\def\Zero{{\mbox{\boldmath $O$}}}

\def\bfc{{\mbox{\boldmath $c$}}}

\def\bff{{\mbox{\boldmath $f$}}}
\def\bfg{{\mbox{\boldmath $g$}}}
\def\bfh{{\mbox{\boldmath $h$}}}

\def\bfm{{\mbox{\boldmath $m$}}}

\def\bfp{{\mbox{\boldmath $p$}}}
\def\bfq{{\mbox{\boldmath $q$}}}
\def\bfr{{\mbox{\boldmath $r$}}}

\def\bfu{{\mbox{\boldmath $u$}}}

\def\bfx{{\mbox{\boldmath $x$}}}

\def\bfD{{\mbox{\boldmath $D$}}}

\def\bfI{{\mbox{\boldmath $I$}}}
\def\bfJ{{\mbox{\boldmath $J$}}}
\def\bfK{{\mbox{\boldmath $K$}}}

\def\bfM{{\mbox{\boldmath $M$}}}

\def\bfQ{{\mbox{\boldmath $Q$}}}

\def\bfSigma{{\mbox{\boldmath $\Sigma$}}}

\def\bfalpha{{\mbox{\boldmath $\alpha$}}}

\def\bfzeta{{\mbox{\boldmath $\zeta$}}}

\def\bfmu{{\mbox{\boldmath $\mu$}}}

\def\bfxi{{\mbox{\boldmath $\xi$}}}

\def\bftau{{\mbox{\boldmath $\tau$}}}

\def\bfphi{{\mbox{\boldmath $\phi$}}}

\def\bfvarphi{{\mbox{\boldmath $\varphi$}}}

%% file: numpsfrags.tex
\psfrag{-100}[cc][][\scalnum]{ $-100$}
\psfrag{-50}[cc][][\scalnum]{ $-50$}
\psfrag{-40}[cc][][\scalnum]{ $-40$}
\psfrag{-30}[cc][][\scalnum]{ $-30$}
\psfrag{-20}[cc][][\scalnum]{ $-20$}
\psfrag{-4}[cc][][\scalnum]{ $-4$}
\psfrag{-2}[cc][][\scalnum]{ $-2$}
\psfrag{-0.4}[cc][][\scalnum]{ $-0.4$}
\psfrag{-0.3}[cc][][\scalnum]{ $-0.3$}
\psfrag{-0.2}[cc][][\scalnum]{ $-0.2$}
\psfrag{-0.1}[cc][][\scalnum]{ $-0.1$}
\psfrag{0}[cc][][\scalnum]{ $0$}
\psfrag{0.002}[cc][][\scalnum]{$0.002$}
\psfrag{0.004}[cc][][\scalnum]{$0.004$}
\psfrag{0.006}[cc][][\scalnum]{$0.006$}
\psfrag{0.008}[cc][][\scalnum]{$0.008$}
\psfrag{0.005}[cc][][\scalnum]{ $0.005$}
\psfrag{0.01}[cc][][\scalnum]{ $0.01$}
\psfrag{0.012}[cc][][\scalnum]{ $0.012$}
\psfrag{0.015}[cc][][\scalnum]{ $0.015$}
\psfrag{0.02}[cc][][\scalnum]{$0.02$}	
\psfrag{0.025}[cc][][\scalnum]{$0.025$}	
\psfrag{0.04}[cc][][\scalnum]{$0.04$}	
\psfrag{0.06}[cc][][\scalnum]{$0.06$}	
\psfrag{0.08}[cc][][\scalnum]{$0.08$}
\psfrag{0.05}[cc][][\scalnum]{ $0.05$}
\psfrag{0.1}[cc][][\scalnum]{ $0.1$}
\psfrag{0.2}[cc][][\scalnum]{ $0.2$}
\psfrag{0.4}[cc][][\scalnum]{ $0.4$}
\psfrag{0.6}[cc][][\scalnum]{ $0.6$}
\psfrag{0.8}[cc][][\scalnum]{ $0.8$}
\psfrag{0.82}[cc][][\scalnum]{ $0.82$}
\psfrag{0.83}[cc][][\scalnum]{ $0.83$}
\psfrag{0.84}[cc][][\scalnum]{ $0.84$}
\psfrag{0.85}[cc][][\scalnum]{ $0.85$}
\psfrag{0.86}[cc][][\scalnum]{ $0.86$}
\psfrag{0.5}[cc][][\scalnum]{ $0.5$}	\psfrag{1}[cc][][\scalnum]{ $1$}
\psfrag{1.5}[cc][][\scalnum]{ $1.5$}
\psfrag{2}[cc][][\scalnum]{ $2$}
\psfrag{2.5}[cc][][\scalnum]{ $2.5$}
\psfrag{3}[cc][][\scalnum]{ $3$}
\psfrag{4}[cc][][\scalnum]{ $4$}
\psfrag{4.5}[cc][][\scalnum]{ $4.5$}
\psfrag{5}[cc][][\scalnum]{ $5$}
\psfrag{5.2}[cc][][\scalnum]{ $5.2$}
\psfrag{5.4}[cc][][\scalnum]{ $5.4$}
\psfrag{5.6}[cc][][\scalnum]{ $5.6$}
\psfrag{5.8}[cc][][\scalnum]{ $5.8$}
\psfrag{6}[cc][][\scalnum]{ $6$}
\psfrag{7}[cc][][\scalnum]{ $7$}
\psfrag{8}[cc][][\scalnum]{ $8$}
\psfrag{9}[cc][][\scalnum]{ $9$}
\psfrag{10}[cc][][\scalnum]{ $10$}
\psfrag{11}[cc][][\scalnum]{ $11$}
\psfrag{12}[cc][][\scalnum]{ $12$}
\psfrag{13}[cc][][\scalnum]{ $13$}
\psfrag{14}[cc][][\scalnum]{ $14$}
\psfrag{15}[cc][][\scalnum]{ $15$}
\psfrag{16}[cc][][\scalnum]{ $16$}
\psfrag{18}[cc][][\scalnum]{ $18$}
\psfrag{20}[cc][][\scalnum]{ $20$}
\psfrag{25}[cc][][\scalnum]{ $25$}
\psfrag{26}[cc][][\scalnum]{ $26$}
\psfrag{28}[cc][][\scalnum]{ $28$}
\psfrag{30}[cc][][\scalnum]{ $30$}
\psfrag{32}[cc][][\scalnum]{ $32$}
\psfrag{34}[cc][][\scalnum]{ $34$}
\psfrag{35}[cc][][\scalnum]{ $35$}
\psfrag{36}[cc][][\scalnum]{ $36$}
\psfrag{38}[cc][][\scalnum]{ $38$}
\psfrag{40}[cc][][\scalnum]{ $40$}
\psfrag{50}[cc][][\scalnum]{ $50$}
\psfrag{60}[cc][][\scalnum]{$60$}
\psfrag{70}[cc][][\scalnum]{$70$}
\psfrag{80}[cc][][\scalnum]{$80$}
\psfrag{90}[cc][][\scalnum]{$90$}
\psfrag{100}[cc][][\scalnum]{$100$}
\psfrag{120}[cc][][\scalnum]{$120$}
\psfrag{140}[cc][][\scalnum]{$140$}
\psfrag{160}[cc][][\scalnum]{$160$}
\psfrag{150}[cc][][\scalnum]{$150$}
\psfrag{180}[cc][][\scalnum]{$180$}
\psfrag{200}[cc][][\scalnum]{$200$}
\psfrag{250}[cc][][\scalnum]{$250$}
\psfrag{300}[cc][][\scalnum]{$300$}
\psfrag{350}[cc][][\scalnum]{$350$}
\psfrag{400}[cc][][\scalnum]{$400$}
\psfrag{500}[cc][][\scalnum]{$500$}
\psfrag{600}[cc][][\scalnum]{$600$}
\psfrag{700}[cc][][\scalnum]{$700$}
\psfrag{800}[cc][][\scalnum]{$800$}
\psfrag{900}[cc][][\scalnum]{$900$}
\psfrag{1000}[cc][][\scalnum]{$1000$}
\psfrag{1200}[cc][][\scalnum]{$1200$}
\psfrag{1400}[cc][][\scalnum]{$1400$}
\psfrag{1600}[cc][][\scalnum]{$1600$}
\psfrag{1800}[cc][][\scalnum]{$1800$}
\psfrag{2000}[cc][][\scalnum]{$2000$}
\psfrag{-0.5}[cc][][\scalnum]{ $\!\!\!-0.5$}
\psfrag{-1}[cc][][\scalnum]{ $\!\!\!-1$}
\psfrag{-1.5}[cc][][\scalnum]{ $\!\!\!-1.5$}
\psfrag{-2.6}[cc][][\scalnum]{ $\!\!\!-2.6$}
\psfrag{-2.8}[cc][][\scalnum]{ $\!\!\!-2.8$}
\psfrag{-3}[cc][][\scalnum]{ $\!\!\!-3$}
\psfrag{-3.2}[cc][][\scalnum]{ $\!\!\!-3.2$}
\psfrag{-5}[cc][][\scalnum]{ $\!\!\!-5$}
\psfrag{-10}[cc][][\scalnum]{ $\!\!\!-10$}

%% file: MainSubmitted.bbl
\begin{thebibliography}{10}

\bibitem{Zanchettin_TRO2013}
B.~Lacevic, P.~Rocco, and A.~M. Zanchettin, ``Safety assessment and control of
  robotic manipulators using danger field,'' {\em IEEE Trans. Robot.}, vol.~29,
  no.~5, pp.~1257--1270, 2013.

\bibitem{Lippi_TCST2020}
M.~{Lippi} and A.~{Marino}, ``Human multi-robot safe interaction: A trajectory
  scaling approach based on safety assessment,'' {\em IEEE Trans. Control Syst.
  Technol.}, pp.~1--16, 2020.

\bibitem{kormushev2011imitation}
P.~Kormushev, S.~Calinon, and D.~G. Caldwell, ``Imitation learning of
  positional and force skills demonstrated via kinesthetic teaching and haptic
  input,'' {\em Advanced Robotics}, vol.~25, no.~5, pp.~581--603, 2011.

\bibitem{Tsumugiwa_ICRA2002}
T.~{Tsumugiwa}, R.~{Yokogawa}, and K.~{Hara}, ``Variable impedance control
  based on estimation of human arm stiffness for human-robot cooperative
  calligraphic task,'' in {\em IEEE Int. Conf. Robot. Autom.}, vol.~1,
  pp.~644--650 vol.1, 2002.

\bibitem{Dimeas_TH2016}
F.~Dimeas and N.~Aspragathos, ``Online stability in human-robot cooperation
  with admittance control,'' {\em IEEE Trans. Haptics}, vol.~9, no.~2,
  pp.~267--278, 2016.

\bibitem{Haddadin_TRO2017}
S.~{Haddadin}, A.~{De Luca}, and A.~{Albu-Schäffer}, ``Robot collisions: A
  survey on detection, isolation, and identification,'' {\em IEEE Trans.
  Robot.}, vol.~33, no.~6, pp.~1292--1312, 2017.

\bibitem{DeLuca_ICRA2005}
A.~D. Luca and R.~Mattone, ``Sensorless robot collision detection and hybrid
  force/motion control,'' in {\em IEEE Int. Conf. Robot. Autom.},
  pp.~999--1004, 2005.

\bibitem{Flacco_ICRA2013}
M.~{Geravand}, F.~{Flacco}, and A.~{De Luca}, ``Human-robot physical
  interaction and collaboration using an industrial robot with a closed control
  architecture,'' in {\em IEEE Int. Conf. Robot. Autom.}, pp.~4000--4007, 2013.

\bibitem{Kouris_RAL2018}
A.~{Kouris}, F.~{Dimeas}, and N.~{Aspragathos}, ``A frequency domain approach
  for contact type distinction in human–robot collaboration,'' {\em IEEE
  Robot. Autom. Lett.}, vol.~3, no.~2, pp.~720--727, 2018.

\bibitem{Cheng_2019}
G.~{Cheng}, E.~{Dean-Leon}, F.~{Bergner}, J.~{Rogelio Guadarrama Olvera},
  Q.~{Leboutet}, and P.~{Mittendorfer}, ``A comprehensive realization of robot
  skin: Sensors, sensing, control, and applications,'' {\em Proceedings of the
  IEEE}, vol.~107, no.~10, pp.~2034--2051, 2019.

\bibitem{Albini_IJRR2020}
A.~Albini and G.~Cannata, ``Pressure distribution classification and
  segmentation of human hands in contact with the robot body,'' {\em Int. J.
  Robot. Res.}, vol.~39, no.~6, pp.~668--687, 2020.

\bibitem{Sharkawy_2018}
A.-N. Sharkawy and N.~Aspragathos, ``Human-robot collision detection based on
  neural networks,'' {\em Int. J. Mechanical Engineering and Robotics
  Research}, vol.~7, no.~2, pp.~150--157, 2018.

\bibitem{sharkawy2020neural}
A.-N. Sharkawy, P.~N. Koustoumpardis, and N.~Aspragathos, ``Neural network
  design for manipulator collision detection based only on the joint position
  sensors,'' {\em Robotica}, vol.~38, no.~10, pp.~1737--1755, 2020.

\bibitem{heo2019collision}
Y.~J. Heo, D.~Kim, W.~Lee, H.~Kim, J.~Park, and W.~K. Chung, ``Collision
  detection for industrial collaborative robots: a deep learning approach,''
  {\em IEEE Robotics and Automation Letters}, vol.~4, no.~2, pp.~740--746,
  2019.

\bibitem{Wahrburg_ECC2019}
N.~{Briquet-Kerestedjian}, A.~{Wahrburg}, M.~{Grossard}, M.~{Makarov}, and
  P.~{Rodriguez-Ayerbe}, ``Using neural networks for classifying human-robot
  contact situations,'' in {\em European Control Conf.}, pp.~3279--3285, 2019.

\bibitem{roman}
M.~{Lippi} and A.~{Marino}, ``Enabling physical human-robot collaboration
  through contact classification and reaction,'' in {\em IEEE Int. Conf. Robot
  and Human Interactive Communication}, pp.~1196--1203, 2020.

\bibitem{hanbook_chapt8}
B.~Siciliano and O.~Khatib, {\em Springer Handbook of Robotics}, ch.~8. Motion
  Control.
\newblock Berlin, Heidelberg: Springer-Verlag, 2016.

\bibitem{BioRob_2012}
A.~{De Luca} and F.~{Flacco}, ``Integrated control for p{HRI}: Collision
  avoidance, detection, reaction and collaboration,'' in {\em IEEE RAS EMBS
  Int. Conf. on Biomedical Robotics and Biomechatronics}, pp.~288--295, 2012.

\bibitem{Redner_EM}
R.~A. Redner and H.~F. Walker, ``Mixture densities, maximum likelihood and the
  {EM} algorithm,'' {\em SIAM Review}, vol.~26, no.~2, pp.~195--239, 1984.

\bibitem{Calinon_IROS20013}
S.~{Calinon}, T.~{Alizadeh}, and D.~G. {Caldwell}, ``On improving the
  extrapolation capability of task-parameterized movement models,'' in {\em
  IEEE/RSJ Int. Conf. on Intelligent Robots and Systems}, pp.~610--616, 2013.

\bibitem{Yuan_Sensors2014}
Q.~Yuan and I.-M. Chen, ``Localization and velocity tracking of human via 3 imu
  sensors,'' {\em Sensors and Actuators A: Physical}, vol.~212, pp.~25 -- 33,
  2014.

\bibitem{deLuca_1992}
A.~D. Luca, G.~Oriolo, and B.~Siciliano, ``Robot redundancy resolution at the
  acceleration level,'' {\em Laboratory Robotics and Automation}, vol.~4,
  pp.~97--106, 1992.

\bibitem{Ames_CDC}
A.~D. {Ames}, J.~W. {Grizzle}, and P.~{Tabuada}, ``Control barrier function
  based quadratic programs with application to adaptive cruise control,'' in
  {\em IEEE Confer. Decis. Control}, pp.~6271--6278, 2014.

\bibitem{basso2020taskpriority}
E.~A. Basso and K.~Y. Pettersen, ``Task-priority control of redundant robotic
  systems using control lyapunov and control barrier function based quadratic
  programs,'' 2020, https://arxiv.org/abs/2001.07547.

\end{thebibliography}
